\documentclass[11pt]{article}
\usepackage{authblk}
\usepackage{paclic34}
\usepackage{algorithm,algcompatible,lipsum}
\usepackage{multicol}
\usepackage{caption}
\usepackage{times}
\usepackage{latexsym}
\usepackage{amsmath}
\usepackage{multirow}
\usepackage{url}

\title{Effective Approach to Develop a Sentiment Annotator For Legal Domain in a Low Resource Setting}

\author[*,1]{Gathika Ratnayaka}
\author[*]{Nisansa de Silva}
\author[*]{Amal Shehan Perera}
\author[+]{Ramesh Pathirana}
\affil[*]{Department of Computer Science and Engineering, University of Moratuwa, Sri Lanka}
\affil[+]{Faculty of Law, University of Colombo, Sri Lanka}
\affil[1]{\texttt{gathika.14@cse.mrt.ac.lk}}

\algnewcommand{\lst}{\texttt{lst}}
\algnewcommand{\slst}{\texttt{slst}}
\algnewcommand{\SEND}{\textbf{send}}

\algdef{SE}[SUBALG]{Indent}{EndIndent}{}{\algorithmicend\ }%
\algtext*{Indent}
\algtext*{EndIndent}

\newsavebox{\algleft}
\newsavebox{\algright}

\savebox{\algleft}{%
\begin{minipage}{.49\textwidth}
  \begin{algorithm}[H]
    \caption{}
    \begin{algorithmic}
      
      \scriptsize
      \STATE \textbf{Function} \(assignSentiment_o(w,sentiment)\)
      
      \Indent
         \IF{sentiment == N} \(D_{on}\cup \{w\}, O_i - \{w\}\)
         \ELSIF{sentiment == P} \(D_{op}\cup \{w\}, O_i - \{w\}\)
         \ENDIF
      \EndIndent    
      \STATE \textbf{EndFunction}
      
      \STATE \textbf{Function} \(assignSentiment_n(w,sentiment)\)
      
      \Indent
         \IF{sentiment == N} \(D_{nn}\cup \{w\}\)
         \ELSIF{sentiment == P} \(D_{np}\cup \{w\}, N_i - \{w\}\)
         \ELSIF{sentiment == O} \(D_{no}\cup \{w\}, N_i - \{w\}\)
         \ENDIF
      \EndIndent    
      \STATE \textbf{EndFunction}
      
      \STATE \textbf{Function} \(assignSentiment_p(w,sentiment)\)
      
      \Indent
         \IF{sentiment == N} \(D_{pn}\cup \{w\},  P_i - \{w\}\)
         \ELSIF{sentiment == P} \(D_{pp}\cup \{w\}\)
         \ELSIF{sentiment == O} \(D_{po}\cup \{w\}, P_i - \{w\}\)
         \ENDIF
      \EndIndent    
      \STATE \textbf{EndFunction}
      \STATE $ P_i=P_m, N_i=N_m, O_i=O_m$, \(D_{on}=\{\},D_{op}=\{\}\)
      \STATE  \(D_{nn},D_{np},D_{no},D_{pp},D_{pn},D_{po}=\{\}\)
      \STATE n=0,p=0 
      \STATE \textbf{While} $1 +|D_{on}| >n$ or $1+|D_{op}|>p $ \textbf{ do}
      \Indent
      \STATE n=$1+|D_{on}|$, p =$1+|D_{op}|$
      \STATE \textbf{for} word w in $O_i$ \textbf{do}
        \Indent
         \STATE l = \(mostSimilar_l(w)\)
         \IF{underRepresented(w) and affinAssignable(w)} \STATE\(assignSentiment_o(w,afinn(w))\)
         \ELSIF{domainSpecific(w) and affinAssignable(w)}
         \STATE\(assignSentiment_o(w,afinn(w))\)
         \ELSIF{domainGeneric(\textit{l}) and $\textit{l}\in N_m \cup D_{on}$}
         \STATE\(\textbf{if } notAntonym(w,l) \textbf{ then } assignSentiment_o(w,N)\)
         \ELSIF{domainGeneric(\textit{l}) and $\textit{l}\in P_m\cup D_{op}$}
          \STATE\(\textbf{if } notAntonym(w,l) \textbf{ then } assignSentiment_o(w,P)\)
         \ENDIF
      \EndIndent  
      \STATE \textbf{end for}
    \EndIndent
      \STATE \textbf{end}
    \end{algorithmic}
  \end{algorithm}%
\end{minipage}}%

\savebox{\algright}{%
\begin{minipage}{.49\textwidth}
  \begin{algorithm}[H]
    \caption{}
    \begin{algorithmic}
      \scriptsize
      \STATE n=0,p=0 
      \STATE \textbf{While} $1 +|D_{nn}| >n$ or $1+|D_{np}|>p $ \textbf{ do}
      \Indent
      \STATE n=$1+|D_{nn}|$, p =$1+|D_{np}|$
      \STATE Q = $N_i \cup D_{on} \cup D_{nn}$, \space R = $P_m \cup D_{op} \cup D_{np}$
      \STATE \textbf{for} word w in $N_i$ \textbf{do}
      \Indent
      \STATE \textit{l} = \(mostSimilar_l(w)\)
      \IF{domainGeneric(w)} \(assignSentiment_n(w,N)\)
      \ELSIF{domainSpecific(w) and affin(w)==N} \STATE \(assignSentiment_n(w,N)\)
      \ELSIF{domainSpecific(w) and notAntonym(w,l)}
      \STATE\textbf{if}{\space$\textit{l} \in Q$} \textbf{ then }
      \space $assignSentiment_n(w,N)$
      \STATE \textbf{else if}  domainGeneric(l) and $l \in R$  \textbf{ then} 
      \Indent
          \STATE \(assignSentiment_n(w,P)\)
      \EndIndent
      \ENDIF
      \EndIndent
      \STATE\textbf{end for}
      \EndIndent
      \STATE\textbf{end}
      \STATE \textbf{for} word w in $N_i$ \textbf{do}
          \(assignSentiment_n(w,O)\) 
      \STATE n=0,p=0 
      \STATE \textbf{While} $1 +|D_{pp}| >p$ or $1+|D_{pn}|>n $ \textbf{ do}
      \Indent
      \STATE p=$1+|D_{pp}|$, n =$1+|D_{pn}|$
      \STATE Q = $N_m \cup D_{on} \cup D_{pn}$, \space R = $P_i \cup D_{op} \cup D_{pp}$
      \STATE \textbf{for} word w in $P_i$ \textbf{do}
      \Indent
      \STATE \textit{l} = \(mostSimilar_l(w)\)
      \IF{domainGeneric(w)} \(assignSentiment_p(w,P)\)
      \ELSIF{domainSpecific(w) and affin(w)==P} \STATE \(assignSentiment_p(w,P)\)
      \ELSIF{domainSpecific(w) and notAntonym(w,l)}
      \STATE\textbf{if}{\space$\textit{l} \in R$} \textbf{ then }
      \space $assignSentiment_p(w,P)$
      \STATE \textbf{else if}  domainGeneric(l) and $l \in Q$  \textbf{ then}
      \Indent
      \STATE $assignSentiment_p(w,N)$
      \EndIndent
      \ENDIF
      \EndIndent
      \STATE \textbf{end for}
      \EndIndent
      \STATE \textbf{end}
      \STATE \textbf{for} word w in $P_i$ \textbf{do}
          \(assignSentiment_p(w,O)\)
      \STATE
      \STATE $P_l = D_{op} \cup D{np} \cup D{pp}$ , \space $N_l = D_{on} \cup D{nn} \cup D{pn}$
    \end{algorithmic}
  \end{algorithm}
\end{minipage}}

\usepackage{amsmath}
\usepackage{tabularx}
\usepackage{diagbox}
\usepackage{multicol}
\usepackage{amsfonts,amssymb}
\usepackage{graphicx}

\usepackage{tabularx}
\usepackage{soul}
\usepackage{epstopdf}
\usepackage[latin1]{inputenc}

\usepackage{hyperref}
\usepackage{xstring}
\usepackage{mathptmx}
\usepackage{graphics}
\usepackage{multirow}

\date{}

\begin{document}
\maketitle
\begin{abstract}
 Analyzing the sentiments of legal opinions available in Legal Opinion Texts can facilitate several use cases such as legal judgement prediction, contradictory statements identification and party-based sentiment analysis. However, the task of developing a legal domain specific sentiment annotator is challenging due to resource constraints such as lack of domain specific labelled data and domain expertise. In this study, we propose novel techniques that can be used to develop a sentiment annotator for the legal domain while minimizing the need for manual annotations of data. 
\end{abstract}

\section{Introduction}

Legal Opinion Texts that elaborate on the incidents, arguments, legal  opinions, and judgements associated with previous court cases are an integral part of case law. As the information that can be acquired from these documents has the potential to be directly applied in similar legal cases, legal officials use of them as information sources to support their arguments and opinions when handling a new legal scenario. Therefore, developing methodologies and tools that can be used to automatically extract valuable information from legal opinion texts while deriving useful insights from the extracted data are of significant importance when it comes to assisting legal officials via automated systems.   

Sentiment analysis can be considered as one such information extraction technique that has a significant potential to facilitate various information extraction tasks. When a legal case is considered, it is built around two major parties that are opposing to each other. The party that brings forward the lawsuit is usually called the plaintiff and the other party is known as the defendant. At the beginning of a legal opinion text, a summary of the case is given, describing the incidents associated with the case and also explaining how each party is related with those incidents. Legal opinions or the opinions of judges about the associated events and laws related to the court case can be considered as the most important type of information available in a legal opinion text. Such opinions may have a positive, neutral, or negative impact on a particular party. In addition to the opinions that are directly related to the conduct of the parties, legal opinion texts also provide interpretations related to previous judgements and also on statutes that are relevant to the legal case. Such opinions may elaborate on the justifications, purposes, drawbacks and loopholes that are associated with a particular statute or a precedent. Moreover, the descriptions also contain information related to the proceeding of court cases such as adjournment of the case and lack of evidence which can be considered as factors that can directly have an impact on the outcomes. When all of the above mentioned factors are considered, sentiment analysis on legal opinion texts can be considered as a task that can facilitate a wide range of use cases. Despite its potential and usefulness, the attempts to perform sentiment analysis in legal domain are limited. This study aims to address this issue by developing a sentiment annotator that can identify sentiments in a given sentence/phrase extracted from legal opinion texts related to the United States Supreme Court. Information that can be derived from such a sentiment annotator can then be adapted to facilitate more downstream tasks such as identifying advantageous and disadvantageous arguments for a particular party, contradictory opinion detection \cite{ratnayaka2019shift}, and predicting outcomes of legal cases \cite{liu2018two} . 

In order to develop a reliable sentiment annotator using supervised learning, it is required to have a large amount of labelled data to train the underlying classification model. However, creating such sophisticated datasets with manually annotated data (by domain experts) for a specialised domain like legal opinion texts is not practical due to extensive resource and time requirements \cite{gamage2018fast,sharma2018identifying}. In a low resource setting, transfer learning can be used as a potential technique to overcome the requirement of creating a sophisticated data set. Adapting these models directly into the legal domain will create drawbacks, especially due to the negative transfer; which is a phenomenon that occurs due to dissimilarities between two domains. Domain specific usage of words, domain specific sentiment polarities and meanings of words can be considered as one major reason that causes negative transfer when adapting datasets/models from one domain to another domain \cite{sharma2018identifying}. In this study, we demonstrate methodologies that can be used to overcome drawbacks due to negative transfer, when adapting a dataset from a source domain to the legal domain. 

\section{Related Work}
\label{related}

It can be observed that the early attempts \cite{thelwall2010sentiment} of developing automatic sentiment analysis mechanisms make use of sentiment lexicons such as AFINN \cite{nielsen2011new}, ANEW \cite{bradley1999affective}, and Sentiwordnet \cite{baccianella2010sentiwordnet}. The sentiment polarity and the strength of a particular word change from one lexicon to another depending on the domain that is being considered when developing the lexicon \cite{nielsen2011new} due to the domain-specific behaviors of words. However, in recent works related to sentiment analysis that are based on machine learning and deep learning techniques, the learning algorithms are allowed to learn the sentiments associated with words and how their compositions affect the overall sentiment of a particular text. The Recursive Neural Tensor Network (RNTN) model proposed by Socher et al. \cite{socher2013recursive} can be considered as an important step towards this direction and it has shown promising results for sentiment classification in movie reviews. However, the performances of such approaches that are based on recursive neural network architectures have been surpassed more recently by the approaches that make use of pretrained language models (eg: BERT\cite{devlin2018bert}), and such approaches have now become the state of the art for sentiment classification \cite{munikar2019fine}. From this point onwards, the RNTN model proposed in \cite{socher2013recursive} will be denoted as $RNTN_m$.  Though the applications of sentiment analysis in the legal domain are limited, there is an emerging interest within the law-tech community to explore how sentiment analysis can be used to facilitate the legal processes\cite{conrad2007opinion,liu2018two}. 
The study by Gamage et al. \cite{gamage2018fast} on performing sentiment analysis in US legal opinion texts can be considered as the closest to our work. However, the direct applicability of their approach into our study is prevented due to some limitations. In \cite{gamage2018fast}, the sentiment annotator was developed to perform a binary-classification task (negative sentiment and non-negative sentiment).  Moreover, one of the key steps in \cite{gamage2018fast} is to identify words that have different sentiments in the legal domain when compared with their sentiments in the movie domain. The identification of such words with domain-specific sentiments had been performed manually by human annotators. However, manually going through a set of words with a significant size is not ideal for a low resource setting in which the intention is to create an optimum outcome from a limited amount of human annotations. Though \cite{sharma2018identifying} proposes an automatic approach based on word embeddings to minimize negative transfer by identifying transferable words that can be used for cross domain sentiment classification, the proposed approach aims only at binary sentiment classification that considers only the positive and negative sentiment classes.

\section{Methodology}

\subsection{Identifying words that can cause negative transfer}

In order to minimize the resource requirements, our intention is to utilize a labeled high resource source domain to facilitate sentiment analysis in a low resource target domain. The Stanford Sentiment Treebank (SST-5) \cite{socher2013recursive} which consists of Rotten Tomato movie reviews labelled according to their sentiments was taken as the source dataset and a corpus of legal opinion texts was selected to extract legal phrases that will be used as the target dataset. As the first step, 3 categories were identified to which the words available in the source dataset can be assigned. The first category is the \textit{Domain Generic words}, the words that behave in a similar manner across the movie review domain and the legal domain. 
The second category is the \textit{Domain Specific words}, the type of words that behaves differently in the two domains and has the potential to cause negative transfer. Within this category, the most frequently used sense/meaning of a word in one domain may differ from that of the other domain. Additionally, such a word may have different sentiment polarities across the two domains.  However, there is another important type of words that can be identified as \textit{Under Represented Words}. The set of \textit{Under Represented Words} consists of words that are frequently occurring in the target domain (legal domain), but are not available or have occurred with a very less frequency in the source dataset. 

Due to the resource limitations, it is not feasible to identify \textit{domain specific words}, \textit{domain generic words}, and \textit{under represented words} manually by going through each word in the legal opinion text corpus. Therefore, the following steps were followed to minimize the requirements for manual annotation. As the first step, stop words in the legal opinion text corpus were removed utilizing the Van stop list \cite{van1979information}. Next \textit{word frequency}, which is the frequency of occurrence of a particular word within the corpus was calculated for each word. Then, the set of words was arranged in a descending order based on the word frequency to create the sorted word set W. From W, first k-words (most frequent k-words) were chosen as the considered set of words S. Here \(k = \min_{j}\{j \in \mathbb{Z}^+|\sum_{i=1}^{j}(w_i) \geq 0.95 \cdot \sum_{i=1}^{n}(w_i)\}\), where \(w_i\) is the \(i^{th}\) element of W and n is the total number of elements in W. 

\usebox{\algleft}

Next, the Stanford Sentiment Annotator ($RNTN_m$) was used to annotate the sentiment of each word in the considered word set S. After the annotation process, the words were distributed into three sets \(P_M\), \(N_M\), \(O_M\) based on the annotated sentiment. The set \(P_M\) is made up of words that were annotated as Very Positive or Positive and the set \(N_M\) is made up of words that were annotated as Very Negative or Negative. The words that were annotated as having a Neutral sentiment were included into \(O_M\). The sets \(P_M\), \(N_M\), \(O_M\) consists of 336, 253, and 4992 words respectively. 
Identifying words in \(O_M\) that have different sentiments across the two domains by manually going through each word is resource extensive as it contains nearly 5000 words. To overcome this challenge and to minimize the required number of manual annotations, we developed a heuristic approach to identify words in the neutral word set (\(O_M\)) that can have different (deviated) sentiments. It should also be noted that in our algorithmic approach, words with deviated sentiments are identified while automatically assigning each word with a legal sentiment (Algorithm 1 and Algorithm 2). 

\usebox{\algright}

Though it is feasible to manually annotate all the words in \(P_M\) and \(N_M\), we have developed our algorithmic approach to identify words that can have deviated sentiments in \(P_M\) and \(N_M\) as well (Algorithm 2) because having a heuristic approach to identify such deviated words can be used to minimize the number of annotations required in case a significant number of words will be identified from \(O_M\) as having deviated sentiments exceeding the annotation budget. Moreover, such an automatic  approach has the potential to be utilized as a mechanism to generate domain specific sentiment lexicons.

 Within our approach to distinguish domain specific words from domain generic words, two key information that can be derived from word embedding models are considered; 1. Cosine similarity between vector representations of two words \textit{u}, \textit{v} as \(Cosine_{domain}(\textit{u}, \textit{v})\) and the most similar word for a particular word \textit{w} as \(mostSimilar_{domain}(\textit{w})\). Domain specific word embeddings have been utilized within our approach to identify domain specific words from domain generic words. The Word2Vec model publicly available at SigmaLaw dataset \cite{sugathadasa2017synergistic} that has been trained using a United States legal opinion text corpus was selected as the legal domain specific word embedding model. The SST-5 dataset does not contain an adequate amount of text data to be used as a corpus to create an effective word embedding model. Therefore, we selected the IMDB movie review corpus \cite{maas2011learning} to train the movie review domain specific Word2Vec embedding model. 
 From this point onwards, \(Cosine_{legal}\) and \(Cosine_{movie-reviews}\) will be denoted by \(Cosine_{l}\) and \(Cosine_{m}\) respectively. Similarly, \(mostSimilar_{legal}(w)\) will be denoted by \(l(w)\) while using \(m(w)\) to denote \(mostSimilar_{movie-reviews}(w)\).  
 
 First, for a given word \textit{w}, we obtain \(l(\textit{w})\) and \(m(\textit{w})\). As Word2Vec \cite{mikolov2013distributed} embeddings are based on distributional similarity, it can be assumed that the most similar word output by a domain specific embedding model to a particular word is related to the domain specific sense of that considered word. For example, \textit{convicted} is obtained as \(l(charged)\). It can be observed that the word convicted is associated with the sense of accusation, which is the most frequent sense of \textit{charge} in the legal domain. However, when it comes to \(m(charged)\), \textit{sympathizing}  is obtained as the output. \textit{Sympathizing} is associated with the sense of \textit{filled with excitement or emotion}, which is the most frequent sense of \textit{charged} in the movie reviews. After obtaining the most similar words for a given word \textit{w}, we define a value $domainSimilarity(w)$ such that \(domainSimilarity(w)\) = \(Cosine_{l}(l(w), m(w))\). As we are considering the legal embedding model when getting the cosine similarity values, a higher \textit{domainSimilarity(w)} value will suggest that legal sense and movie sense of the word \textit{w} have a similar meaning in the legal domain while a lower \textit{domainSimilarity(w)} will suggest that the meanings of the two senses are less similar to each other. For example, the value obtained for \textit{domainSimilarity(Charged)} was 0.06 while it was 0.53 for \textit{domainSimilarity(Convicted)} (\textit{convicted} has a similar sense across the two domains).

The next step is to identify a threshold based on domainSimilarity(w) to heuristically distinguish whether a word \textit{w} is domain generic or not. To that regard, we made use of already available Verb Similarity dataset \footnote{https://osf.io/bce9f/} developed for the legal domain. The dataset consists of 959 verb pairs manually annotated based on whether the two verbs in a pair have a similar meaning or not. First, a threshold \textit{t} based on cosine similarity was defined. For a given two verbs $v_i,v_j$, if \(Cosine_{l}(v_i, v_j)\) $\geq$ \textit{t}, the two verbs are considered as having a similar meaning.  From the experiments, it was observed that precision is less than 0.5 when the threshold value is equal to 0.1. Therefore, 0.2 is selected as the threshold value to identify domain generic words based on the \textit{domainSimilarity(w)} score. In other words, if \textit{domainSimilarity(w)} is greater than or equal to 0.2, the word \textit{w} will be considered as domain generic and the attribute \textit{domainGeneric(w)} will be set to true. Otherwise, the attribute \textit{domainSpecific(w)} will be set to true. Though we have used the aforementioned approach to determine the threshold, it is a heuristic and domain specific value that can be decided based on different experimental techniques (when applying this methodology to another domain). 

Even if a word behaves in a similar manner across the two domain, it still can be assigned with a wrong sentiment (neutral sentiment) due to under representation.  However, it is important to identify words with sentiment polarities (positive or negative) as the descriptions with positive or negative sentiments tend to contain more specific information that will be useful in legal analysis. As a measure of identifying sentiment polarities of under represented words, we made use of AFINN \cite{nielsen2011new} sentiment lexicon (denoted as set A from this point onwards), which consists of 3352 words annotated based on their sentiment polarity (positive, neutral, negative) and sentiment strength considering the domain of twitter discussions.  If a frequency of a word \textit{w} is less than 3 in the source dataset, \textit{underRepresented(w)} is set to true. Assignment of AFINN sentiment for an under represented word or a domain specific word \textit{w} can create a positive impact if the most frequently used sense of \textit{w} in twitter discusion domain is aligned towards it's sense in the legal domain than the sense of that word (w) in the movie review domain. In order to heuristically determine this factor, we have defined an attribute name \textit{afinnSimilarity} such that  \(afinnSimilarity(w)\) = \(Cosine_{t}(w, l(w)) - Cosine_{t}(w, m(w))\), where w is a given word and \(Cosine_{t}\) is the cosine similarity obtained using a publicly available Word2Vec model \cite{godin2019} trained using tweets.  If  \(Cosine_{t}(w, l(w))\) $>$  \(Cosine_{t}(w, m(w))\), it can be assumed that the sense of word \textit{w} in twitter discussions is more closer to its sense in the legal domain than that of the movie-reviews. Thus, if  \textit{afinnSimilarity(w)} $>$ 0 and $w \in A$, the attribute \textit{afinnAssignable(w)} is set to true.

Both Algorithm 1 and Algorithm 2 are two parts of one major algorithmic approach (Algorithm 1 executes first). Therefore, the functions and attributes defined in Algorithm 1 are applied globally for both Algorithm 1 and Algorithm 2 and the states of the attributes after executing Algorithm 1 will be transferred to the Algorithm 2. In the algorithms, P, N, O denotes positive, negative, and neutral sentiments respectively. \textit{afinn(w)} is the AFINN sentiment categorization of a given word w.
When observing the algorithm, it can be observed that sentiment of \(l(w))\) is also considered when determining the correct sentiments of a word. For a word in $O_m$, the sentiment of \(l(w)\) will be assigned if \(l(w)\) is \textit{domain generic} (Algorithm 1). This step was followed as another way to identify words with sentiment polarities (positive or negative). The sentiments of domain generic words in $P_m$ or $N_m$ will not be changed under any condition. For a domain specific word w in $P_m$ or $N_m$, if \(l(w)\) has a opposite sentiment polarity to that of w, the sentiment of \(l(w)\) will be assigned to w only if \(l(w)\) is domain generic. 
All the \textit{domain specific} words in $P_m$ or $N_m$ that do not satisfy any of the conditions that are required to assign a positive or negative polarity (Algorithm 2), will be assigned with a neutral sentiment. This step is taken because such \textit{domain specific} words have a relatively higher probability to have opposite sentiment polarities in the legal domain, thus capable of transferring wrong information to the classification models \cite{sharma2018identifying}. 
Assigning neutral sentiment will reduce the impact of negative transfer that can be caused by such words (neutral sentiment is better than having the opposite sentiment polarity). Furthermore, it should be noted that an antonym of a particular word \textit{w} can be given as \(l(w)\) by the embedding model due to semantic drift. To tackle this challenge, WordNet \cite{fellbaum2012wordnet} was used to check whether a given word \textit{w} and \(l(w)\) are antonyms. 
If they are not antonyms, notAntonyms() attribute is set true. After running the Algorithm 1 and 2 by taking $P_m, O_m, N_m$ as the inputs, the word sets $D_{on}, D_{op}$ were obtained that consist of words the overall algorithm picked from $O_m$ as having negative and positive sentiments respectively. $D_{on}, D_{op}$ together with $P_m, N_m$ were given to a legal expert in order to annotate the words in these sets based on their sentiments. 
$|D_{on}|$ = 220 and $|D_{op}|$=116, thus reducing the required amount of annotations to 925 (925= $|W|$, where $W = D_{op}\cup D_{on}\cup P_m \cup N_m $). After the annotation process, three word sets $N_a,O_a,P_a$ were obtained that contains words that are annotated as having positive, neutral and negative sentiments respectively. 
Then word sets $D_n, D_o, D_p$ were created such that $D_n=\{w \in W|w \in N_a \& w \notin N_m\}$,
$D_p=\{w \in W|w \in P_a \& w \notin P_m\}$, $D_o=\{w \in W|w \in O_a \& w \notin O_m$\}. 
$P_l$ contains the set of words identified by the overall algorithm as having positive sentiment and $N_l$ contains the words identified as having negative sentiment (without human intervention).

\subsection{Fine Tuning the RNTN Model}
As an approach to develop a sentiment classifier for legal opinion texts,  $RNTN_m$ (Stanford Sentiment Annotator) \cite{socher2013recursive} was fine tuned following a similar methodology as proposed by \cite{gamage2018fast}. In the proposed methodology \cite{gamage2018fast}, there is no need to further train the $RNTN_m$ model or to modify the neural tensor layer of the model. Instead, the approach is purely based on replacing the word vectors. In this approach, if a word \textit{v} in a word sequence \textit{S} have a deviated sentiment $s_d$ in the legal domain when compared with its sentiment $s_m$ as output by the $RNTN_m$, the vector corresponding to \textit{v} will be replaced by the vector of word \textit{u}, where \textit{u} is a word from a list of predefined words that has the sentiment $s_d$ as output by $RNTN_m$. When choosing \textit{u} from the list of predefined words, PoS tag of \textit{w} in word sequence \textit{S} is considered in order to preserve the syntactic properties of the language. For example, if we consider the phrase \textit{Sam is charged for a crime}, as \textit{charged} is a word that have a deviated sentiment, the vector corresponding to \textit{charged} will be substituted by the vector of \textit{hated} (\textit{hated} is the word that matches the PoS of \textit{charged} from the predefined word list corresponding to the negative class) \cite{gamage2018fast}. When extending the approach proposed in \cite{gamage2018fast} for three class sentiment classification, a predefined word list for positive class was developed by mapping a set of selected words that have positive sentiment in $RNTN_m$ to each PoS tag. The mapping can be represented as a dictionary R, where R = \{JJ:beautiful, JJR:better, JJS:best, NN:masterpiece, NNS:masterpieces, RB:beautifully, RBR:beautifully, RBS:beautifully, VB:reward, VBZ:appreciates, VBP:reward, VBD:won, VBN:won, VBG:pleasing\}. For the negative class and the neutral class, the PoS-word mappings provided by \cite{gamage2018fast} for negative and non-negative classes were used respectively. Furthermore, instead of annotating each word in the selected vocabulary to identify words with deviated sentiments, we used  word sets $D_n,D_o,D_p$ that were derived using the approaches described in Section 3.1. In Section 4, the fine tuned RNTN model developed in this study is denoted as $RNTN_l$.

\subsection{Adapting the BERT based approaches}

An approach based on $BERT_{large}$ embeddings \cite{munikar2019fine} has achieved the state of the art results for sentiment classification of sentences in SST-5 dataset. In order to adapt the same approach for our task, following steps were followed. First, sentences with their sentiment labels were extracted from SST-5 training set. The SST-5 training set consists of 8544 sentences labelled for 5 class sentiment classification. As our focus is on 3 class classification, the sentiment labels in the SST training set were converted for 3 class sentiment classification by mapping very positive, positive labels as positive and very negative,negative labels as negative. Next, following a similar methodology as described in \cite{munikar2019fine}, \textit{canonicalization}, \textit{tokenization} and \textit{special token addition} were performed as the preprocessing steps. Then, the classification model was designed following the same model architecture described in \cite{munikar2019fine}, that consists of a dropout regularization and a softmax classification layer on top of the pretrained BERT layer. Similarly to \cite{munikar2019fine}, $BERT_{large}$ uncased was used as the pretrained model and during the training phase, dropout of probability factor 0.1 was applied as a measure of preventing overfitting. Cross Entropy Loss was used as the cost funtion and stochastic gradient descent was used as the optimizer (batch size was 8). Then, the model was trained using the SST-5 training sentences. As information related to number of training epoch could not be found in \cite{munikar2019fine}, we experimented with 2 and 3 epcohs and calculated the accuracies with a test set of 500 legal phrases (Section 4). When trained for 2 epochs, the accuracy was 57\% and for 3 epcohs it was reduced to 52\%, possibly due to the overfitting with the source data. Therefore, 2 was choosen as the number of training epochs. This model will be denoted as $BERT_m$ in next sections.

In order to finetune the BERT based approach to the legal sentiment classification, the following steps were followed. First we selected sentences in the SST training data, that consists of words that were identified as having deviated sentiments (words in $D_o\cup D_p \cup D_n $). If the sentiment label of the sentence S that has a deviated sentiment word w is different from the sentiment label assigned to w by the legal expert, then S will be removed from the original SST training dataset as a measure of reducing negative transfer. For example, if there is a sentence S with word \textit{charged} and if the sentiment of S is positive or neutral (sentiment of charged is negative in legal domain), then that sentence S will be removed from the training set. After removing such sentences, the training set was reduced to 6318 instances and this new training set will be denoted by D from this point forward. Next, for each word \textit{w} in $D_n$ or $D_p$, we randomly selected 2 sentences that contains \textit{w} from the legal opinion text corpus. Then, the sentiments of the selected sentences were manually annotated by a legal expert. As $|D_n| = 206$ and $|D_p| = 82$, only 576 new annotations were needed ($|D_o| = 230$, but words in $D_o$ were not considered for this approach as they are having a neutral legal sentiment). Then, these 576 sentences from legal opinion texts were combined together with sentences in D, thus creating a new training set L that consists of 6894 instances. The above mentioned steps were followed to remove the negative transfer from the source dataset and also to fine tune the dataset to the legal domain. Then, L was used to train a BERT based model using the same architecture, hyper parameters and number of training epochs that were used to train $BERT_m$. The model obtained after this training process is denoted as $BERT_l$.  

\small\addtolength{\tabcolsep}{0 pt}
% \usepackage{multirow}
\begin{table*}[]

\caption{Evaluating the word lists generated from Algorithm 1 and Algorithm 2}
\label{table:lexicon}
\centering
\begin{tabular}{|l|l|l|l|c|c|c|c|c|c|c|}
\hline
\multirow{2}{*}{\backslashbox{Polarity}{Metric}} & \multicolumn{4}{c|}{Number of Words} & \multicolumn{4}{c|}{Percentages}   \\ \cline{2-9} 
                  & {$N_m$}           & {$N_l$}            & {$P_m$}           & {$P_l$} & {$N_m$}           & {$N_l$}            & {$P_m$}           & {$P_l$}                                       \\ \hline \hline
Negative       & \textbf{154}        & \textbf{317}        & 17        & 20        & \textbf{61\%}        & \textbf{80\%}       & {5\%}        & {7\%}                              \\ \hline
Neutral      & 96        & 73        & 180        & 89        & {38\%}        & {19\%}         & {54\%}        & {41\%}                             \\ \hline
Positive      & 3        & 4        & \textbf{139}        & \textbf{181}        & {1\%}       & {1\%}        & \textbf{41\%}        & \textbf{62\%}                         \\ \hline
Total     & 253        & 394        & 336        & 290        & {100\%}        & {100\%}        & {100\%}       & {100\%}         \\ \hline

\end{tabular}
\end{table*}

\section{Experiments and Results}
\subsection{Identification of words with deviated sentiments}
In order to evaluate the effectiveness of the proposed algorithmic approach when it comes to identifying legal sentiment of a word, we have compared the positive word list ($P_l$) and negative word list($N_l$) identified by the algorithm with $P_m$ and $N_m$ respectively as shown in Table 1. The way in which $P_l$ and $N_l$ were obtained is described in Algorithm 2. It can be observed that the precision of identifying words with negative sentiments is 80\% in the algorithmic approach and it is a 19\% improvement when compared with the $RNTN_m$ \cite{socher2013recursive}. Furthermore, the number of correctly identified negative words have increase to 317 from 154. Though the precision of identifying words with positive sentiment is only 62\%, there is an improvement of 21\% when compared with the $RNTN_m$. Precision of identifying words with positive sentiment is relatively low due to the fact that most of the words that have a positive sentiment in generic language usage have a neutral sentiment in the legal domain. Sophisticated analysis in relation to the neutral class could not be performed due to the large amount of words available in $O_m$. When considering these results, it can be seen that the proposed algorithm has shown promising results when it comes to determining the legal domain specific sentiment of a word. Additionally, it implies that the proposed algorithmic approach is successful in identifying words that have different sentiments across the two domains. This approach can also be extended to other domains easily as domain specific word embedding models can be trained using an unlabelled corpus. Furthermore, the proposed algorithmic approach also has the potential to be used in automatic generation of domain specific sentiment lexicons.

\small\addtolength{\tabcolsep}{0 pt}
% \usepackage{multirow}
\begin{table*}[]

\caption{Precision(P), Recall (R) and F-Measure (F) obtained from the considered models}
\label{table:bert_models}
\centering
\begin{tabular}{|l|l|l|l|c|c|c|c|c|c|c|c|c}
\hline
\multirow{2}{*}{\backslashbox{Model}{Metric}} & \multicolumn{3}{c|}{Negative} & \multicolumn{3}{c|}{Neutral} & \multicolumn{3}{c|}{Positive} &  \multirow{2}{*} {Accuracy} \\ \cline{2-10} 
                  & P           & R           & F           & P           & R           & F           & P           & R           & F                               \\ \hline \hline
$RNTN_m$       & 0.51        & 0.68        & 0.58        & 0.44        & 0.52        & 0.48       & 0.48        & 0.10        & 0.16 & 0.48                      \\ \hline
$RNTN_l$ (Improved)      & 0.55        & 0.70        & 0.62        & 0.54        & 0.51        & 0.52        & 0.73        & 0.44        & 0.55         & 0.57              \\ \hline
$BERT_m$      & 0.68        & 0.73        & 0.70        & 0.47        & 0.68        & 0.56        & 0.57        & 0.13        & 0.21 & 0.57                    \\ \hline
$BERT_l$ (Improved)     & 0.72        & 0.79        & 0.75        & 0.58        & 0.55        & 0.57        & 0.70        & 0.62        & 0.66 & 0.67                       \\ \hline
\end{tabular}
\end{table*}

\subsection{Sentiment Classification}
In order to evaluate the performances of the considered models when it comes to legal sentiment classification, it is needed to prepare a test set that consists of sentences from legal opinion texts annotated according to their sentiment. As the first step of preparing the test set, 500 sentences were randomly picked from the legal opinion text corpus such that there is no overlap between the test set and the sentences used to train $BERT_l$. Then sentiment of each sentence was annotated by a legal expert. According to the human annotations, the number of data instances belong to negative, neutral and positive classes in the test set were 211, 168, and 121 respectively. The results obtained for each model for the test set is shown in Table 2. The effectiveness of the fine tuning approaches proposed in this study is evident as the RNTN finetuning has achieved accuracy increase of 9\% while fine tuning the dataset for BERT training has achieved an accuracy increase of 10\% when compared with the performances of the respective source models. It can be observed that the $BERT_m$ has the same accuracy as the $RNTN_l$. However, the performance of $RNTN_l$ model is relatively consistent across all 3 classes while the recall, f-measure of $BERT_m$ in relation to the positive class is significantly low. It should be noted that $BERT_l$ model that was trained after fine tuning the dataset for legal domain outperforms all other models. Furthermore, the state of the art accuracy value for 5 class sentiment classification of sentences in SST-5 dataset is 55.5\%\cite{munikar2019fine}. An accuracy of 67\% for 3 class classification in the legal domain can be considered as  satisfactory when we consider the added language complexities in legal opinion texts, though the number of classes has been reduced to 3. Most importantly, the accuracy enhancement of 10\% compared with $BERT_m$ was achieved by including only 576 new sentences from legal opinion texts that were annotated by a legal expert. Therefore, it can be concluded that the transfer learning approach mentioned in Section 3.3 is an effective way to develop a domain specific sentiment annotator with a considerable accuracy while utilizing a minimum amount of annotations.

\section{Conclusion}

Developing a sentiment annotator to analyze the sentiments of legal opinions can be considered as the primary contribution of this study. In order to achieve this primary objective in a low resource setting, we have proposed effective approaches based on transfer learning while utilizing domain specific word representations to overcome negative transfer. As a part of the overall methodology, we have also proposed an algorithmic approach that has the capability of identifying the words with deviated sentiments across the source and target domains, while assigning the target domain specific sentiment to the considered words. The data sets prepared within this study for testing and training purposes has been made publicly available \footnote{https://osf.io/zwhm8/}. Moreover, the methodologies formulated in this study are designed in a way such that they can be easily adaptable for any other domain.

\section*{Acknowledgments}

This research was funded by SRC/LT/2018/08 grant of University of Moratuwa.

\clearpage
\bibliography{references} 
\bibliographystyle{acl}

\end{document}